\documentclass[letterpaper, 10 pt, journal, twoside]{IEEEtran}  

\usepackage{amsthm}
\usepackage{amsmath}
\usepackage{amssymb}
\usepackage{mathtools} 
\usepackage{xcolor}

\usepackage{hyperref}
\usepackage{bm}
\newcommand{\ssymbol}[1]{^{\@fnsymbol{#1}}}

\newcommand{\infnorm}[1]{\left\Vert#1\right\Vert_{\infty}}
\newcommand{\twonorm}[1]{\left\Vert#1\right\Vert_2}

\newcommand{\jw}{(j\omega)}
\newcommand{\abs}[1]{\left\lvert#1\right\rvert}
\newcommand{\omegasample}{\omega_\text{sample}}
\DeclareMathOperator*{\argmax}{arg\,max}

\usepackage{graphicx}
\usepackage{subfig}
\usepackage{booktabs}
\usepackage{multirow}
\usepackage{multicol}
\usepackage{siunitx}

\newtheorem{theorem}{Theorem}

\newtheorem{remark}{Remark}

\newtheorem{example}{Example}

\usepackage{algpseudocode}
\usepackage[ruled,vlined,linesnumbered]{algorithm2e}
\SetKwInput{Input}{Input}
\SetKwInput{Output}{Output}
\SetKwInput{Init}{Initialize}

\algnewcommand\algorithmicinput{\textbf{Input:}}
\algnewcommand\Input{\item[\algorithmicinput]}
\algnewcommand\algorithmicoutput{\textbf{Output:}}
\algnewcommand\Output{\item[\algorithmicoutput]}
\algnewcommand\algorithmicinit{\textbf{Initialize:}}
\algnewcommand\Init{\item[\algorithmicinit]}
\makeatother
\usepackage{cite}
\usepackage[skip=8pt,font=footnotesize]{caption}
\usepackage{bbm} 

\newcommand{\diffhighlight}[1]{\textcolor{black}{#1}}
\newcommand{\diffht}[1]{\textcolor{black}{#1}}
\makeatletter
\def\namedlabel#1#2{\begingroup
    #2%
    \def\@currentlabel{#2}%
    \phantomsection\label{#1}\endgroup
}

\title{To Share or Not to Share? Performance Guarantees and the Asymmetric Nature of Cross-Robot Experience Transfer
}

\author{Michael J. Sorocky, Siqi Zhou, and Angela P. Schoellig
\thanks{The authors are with the Dynamic Systems Lab (\href{http://www.dynsyslab.org}{www.dynsyslab.org}), Institute for Aerospace Studies, University of Toronto, Canada and the Vector Institute for Artificial Intelligence, Toronto. Emails: \{msorocky, siqi.zhou\}@robotics.utias.utoronto.ca, schoellig@utias.utoronto.ca}%
}

\begin{document}

\maketitle
\thispagestyle{empty}
\pagestyle{empty}

\begin{abstract}

In the robotics literature, experience transfer has been proposed in different learning-based control frameworks to minimize the costs and risks associated with training robots. While various works have shown the feasibility of transferring prior experience from a source robot to improve or accelerate the learning of a target robot, there are usually no guarantees that experience transfer improves the performance of the target robot. In practice, the efficacy of transferring experience is often not known until it is tested on physical robots. This trial-and-error approach can be extremely unsafe and inefficient. Building on our previous work, in this paper we consider an inverse module transfer learning framework, where the inverse module of a source robot system is transferred to a target robot system to improve its tracking performance on arbitrary trajectories. We derive a theoretical bound on the tracking error when a source inverse module is transferred to the target robot and propose a Bayesian-optimization-based algorithm to estimate this bound from data. We further highlight the asymmetric nature of cross-robot experience transfer that has often been neglected in the literature. We demonstrate our approach in quadrotor experiments and show that we can guarantee positive transfer on the target robot for tracking random periodic trajectories.

\end{abstract}
\section{Introduction}
 \IEEEPARstart{A}{pproaches} to transfer experience across robots or across tasks have been explored in different robot learning-based control frameworks (e.g., \cite{hamer2013knowledge,bocsi2013alignment,chowdhary2013rapid,pereida2018data}). The goal of experience transfer is to leverage existing data or a learned model to accelerate or improve the learning process on a new robot or new task~\cite{bocsi2013alignment}. While existing literature shows the feasibility to improve or accelerate the learning of a robot by leveraging prior experience in training, it is often implicitly assumed without further analysis that the transferred experience will lead to improved performance (i.e., positive transfer) on the target robot or task. This assumption is typically not examined until it is experimentally tested on the target robot hardware. This trial-and-error experience transfer strategy can be unsafe and inefficient for practical robot applications.
 
 \begin{figure}
    \centering
    \subfloat[Inverse module transfer learning framework]{%
        \includegraphics[width=\columnwidth]{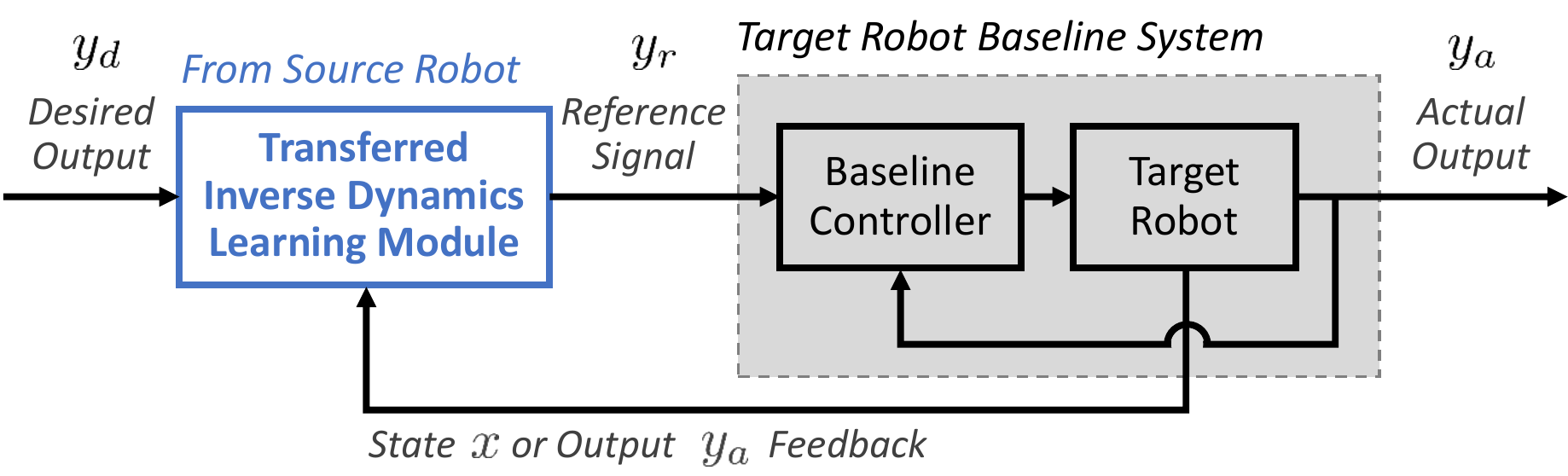}
        \label{subfig:block-diagram}
    }%
    \hfill%
    \subfloat[Source and target quadrotor platforms used in experiments
    ]{%
        \includegraphics[width=\columnwidth]{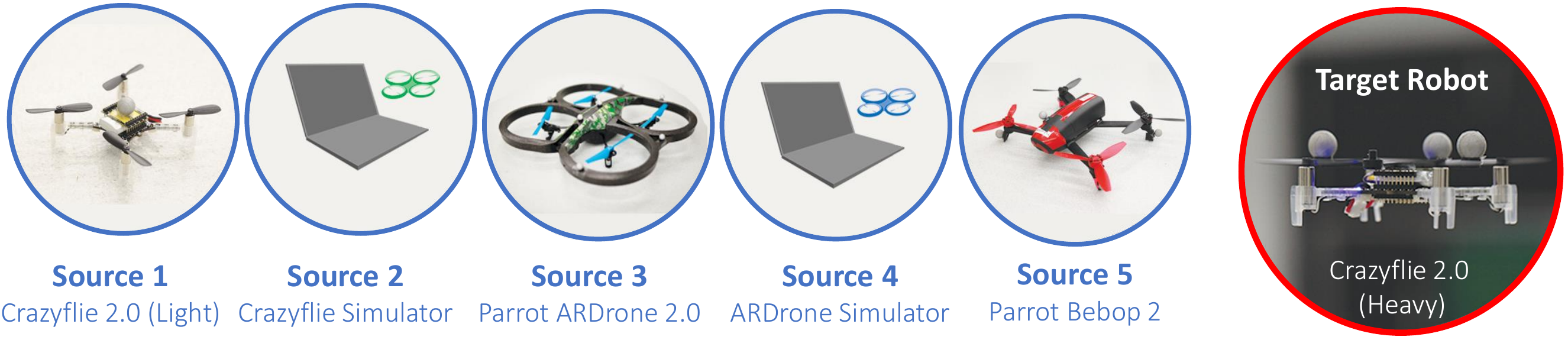}
        \label{subfig:source-target-quadrotors}
    }%
    \vspace{0.5em}
    \caption{The inverse learning control architecture~\cite{zhou-cdc17} considered in our experience transfer analysis is shown in~(a). The goal is to enhance the performance of the target robot system (grey block) for tracking arbitrary trajectories. The blue block represents the inverse dynamics learning module transferred from a source robot system. The transferred inverse block adjusts the reference signal $y_r$ sent to the target robot baseline system such that the map from the desired output $y_d$ to the actual target system output $y_a$ is closer to identity~\cite{zhou-cdc17}. In this paper, we aim to provide performance guarantees for this transfer approach \textit{prior} to testing source inverse modules on the target robot. We demonstrate our approach in quadrotor experiments on the source and target quadrotors shown in~(b). A video of this work can be found here: \url{http://tiny.cc/transfer_guarantees}}
    \label{fig:block-diagram}
    \vspace{-0.5em}
\end{figure}

In this paper, building on our previous work~\cite{zhou-cdc17}, we consider an inverse module transfer learning framework, where we transfer an inverse dynamics module previously trained on a source robot to improve the tracking performance of a target robot. \diffhighlight{While classical approaches such as model predictive control (MPC), the linear quadratic regulator (LQR), or PID control can be applied to trajectory tracking problems, these approaches typically rely on a sufficiently accurate dynamics model of the robot or can be time consuming to tune for achieving high-performance tracking on arbitrary trajectories. With the inverse dynamics learning framework~\cite{zhou-cdc17}, we can improve the tracking performance of a nonlinear control-affine robot system whose dynamics are not exactly known.} As shown in Fig.~\ref{subfig:block-diagram}, the learned inverse module (blue) is pre-cascaded to the target robot system (grey), with the goal of achieving an identity map between the desired output $y_d$ and the actual output $y_a$. In~\cite{li-icra17,zhou-cdc17}, we show that we can train a deep neural network (DNN) to approximate the inverse dynamics of a source quadrotor's baseline control system, which reduced the tracking error of the source quadrotor by an average of 43\% on arbitrary hand-drawn trajectories.

Our goal in this paper is to provide performance guarantees when a source inverse module is transferred to a target robot. With this inverse module transfer, we aim to reduce the time and data to train the target robot while guaranteeing improved performance. Our contributions are as follows:
\begin{enumerate}
    \item derive theoretical bounds on the tracking error when a source inverse module is transferred to the target robot,
    \item introduce a Bayesian Optimization (BO)-based algorithm to estimate the tracking error upper bound from simple periodic experiments on the source and target robots for guaranteeing positive transfer,
    \item highlight the asymmetric nature of this inverse transfer learning framework, and
    \item demonstrate our approach in quadrotor experiments.
\end{enumerate}
  
\section{Related work}
The idea of experience transfer or transfer learning originates from machine learning research. It has been applied to fields such as computer vision~\cite{shao2014transfer}, natural language processing~\cite{wang2015transfer}, and reinforcement learning~\cite{taylor2009transfer}. The goal of transfer learning is to develop methods that exploit information from source domains to facilitate learning in a target domain. 

Transfer learning has been considered in different robot learning-based control frameworks. 
Experience transfer approaches in control can be categorized as cross-task transfer or cross-robot transfer \cite{hamer2013knowledge,chowdhary2013rapid,pereida2018data,bocsi2013alignment}. Whether it is robot model learning~\cite{bocsi2013alignment}, controller learning~\cite{chowdhary2013rapid}, or reference adaptation~\cite{hamer2013knowledge,pereida2018data}, the common goal of transferring experience is to leverage prior knowledge such that the time and data required for training a new robot or on a new task can be reduced. Techniques from adaptive control are used in~\cite{chowdhary2013rapid} to transfer controller parameters from a source robot to a target robot.  In~\cite{bocsi2013alignment}, data collected from a source robot is mapped to align with data from the target robot to speed the learning of the forward kinematics of a target manipulator. In~\cite{helwa2017multi}, the authors consider a similar alignment-based framework and show that the optimal alignment map is a dynamical system.

While existing work demonstrates transferring prior experience can facilitate learning on new robots or tasks, there is usually no guarantee that using prior experience will improve the performance of the target robot before testing. The phenomenon of the performance of the target robot being degraded when using experience from a source robot is known as \textit{negative transfer}. As demonstrated in machine learning literature~\cite{wang2019characterizing,rosenstein2005transfer}, negative transfer can occur when the source and target domains are dissimilar. In our previous work~\cite{sorocky-icra20}, we also showed that using experience from a dissimilar source robot can cause negative transfer.

In this work, we consider a cross-robot inverse module transfer problem and derive a procedure that allows us to guarantee positive transfer prior to transferring experience from the source robot to the target robot.

\section{Problem Formulation}
   In this paper, we consider the control architecture shown in Fig.~\ref{subfig:block-diagram}, where an inverse dynamics module from a source robot~$R_s$ is transferred to enhance the tracking performance of a target robot~$R_t$. We say that the transfer is \textit{positive} if the performance of the target robot with the source robot inverse is improved as compared to the target robot baseline system. Our goal is to provide a systematic approach that allows us to identify when we achieve positive transfer from the source robot to the target robot prior to physically testing the source inverse module on the target robot. 
   In our analysis, we make the following assumptions:
    \begin{enumerate}
        \item[\text{\namedlabel{assumption:min-phase}{(A1)}}] The baseline systems of the source and the target robots are stable and have stable inverse dynamics. 
        \item[\text{\namedlabel{assumption:linear-siso}{(A2)}}] The source and target baseline systems are single-input single-output (SISO), linear time-invariant (LTI) systems, where the relation between the reference signal~$y_r(t)$ and the system output~$y_a(t)$ can be represented by proper transfer functions of the form $G(s) = \frac{\hat{y}_a(s)}{\hat{y}_r(s)}$ with $\hat{y}_r(s)$ and $\hat{y}_a(s)$ denoting the Laplace transforms of $y_r(t)$ and $y_a(t)$, respectively.
        \item[\text{\namedlabel{assumption:yd-bounded-2norm}{(A3)}}] The desired trajectory $y_d$ satisfies $\twonorm{y_d} < \infty$, where $\twonorm{\cdot}$ denotes the $l_2$-norm~\cite{doyle2013feedback}. 
    \end{enumerate}
    Note that assumption \ref{assumption:min-phase} is a necessary condition for safely applying the inverse dynamics learning approach to improve the tracking performance of a black-box system~\cite{zhou-cdc17}. In \ref{assumption:linear-siso}, we assume the baseline systems of the source and the target robots are SISO, LTI systems to simplify our analysis. \diffhighlight{For many practical robot systems, we can use linearization techniques in baseline tracking controller designs. Applying these techniques to quadrotors~\cite{helwa2018construction}, fully-actuated manipulators~\cite{spong2006robot} or wheeled ground robots~\cite{giesbrecht2009vision} typically results in decoupled linear dynamics. The decoupling resulting from linearization allows us to apply our analysis to each SISO system separately and provide insights on transfer performance for a range of practical systems~(e.g., \cite{helwa2018construction,spong2006robot,giesbrecht2009vision}).}
    Lastly, assuming the desired trajectory has finite $l_2$-norm \ref{assumption:yd-bounded-2norm} is not restrictive, and can be satisfied with a desired trajectory that is bounded, and is non-zero over a finite-time horizon $[0, T]$ with $T \mathord{<} \infty$. 
    
    In our discussion, we denote the transfer functions of the source and target robots by $G_s(s)$ and $G_t(s)$, respectively. From~\ref{assumption:min-phase}, we assume that $G_s(s)$ and $G_t(s)$ and their inverses are bounded-input and bounded-output (BIBO) stable. The tracking error of the target system is defined as ${e(t) = y_a(t) - y_d(t)}$. We consider two cases: \textit{(i)}  using an inverse module from a source robot, in which case we denote the tracking error by $e_{t,s}$, and \textit{(ii)} using the baseline target system without an inverse module (i.e., setting ${y_r = y_d}$), in which case we denote the tracking error by $e_{t,b}$. We say that \textit{positive transfer} occurs when using an inverse module from a source robot if $\twonorm{e_{t,s}} < \twonorm{e_{t,b}}$.

\section{Main Results}
    \label{sec:main-results}

    To ensure safe operation when using an inverse module from a particular source system $R_s$ for the target $R_t$, our first goal is to find bounds on the corresponding tracking error $\twonorm{e_{t,s}}$ \textit{before} testing on the target robot as shown in Fig.~\ref{subfig:block-diagram}. In Sec.~\ref{subsec:error-bounds}, we derive an upper bound on the tracking error $\twonorm{e_{t,s}}$ and relate the bound to our goal of identifying positive transfer. In Sec.~\ref{subsec:bo-algorithm}, we propose an algorithm to estimate the bound from data when the source and target dynamics are unknown. In Sec.~\ref{subsec:asymmetry}, we demonstrate the asymmetric nature of transfer learning in our framework.

    \subsection{Bound on the Tracking Error} 
        \label{subsec:error-bounds}
        
        In this subsection, we assume that the dynamics of the source and target robot systems are known and provide an analysis of the tracking error $e_{t,s}$ of the target robot when the transferred source robot inverse module is applied. \diffhighlight{We derive an upper bound on $\twonorm{e_{t,s}}$ using the signal norm analysis from \cite{doyle2013feedback}, which relates the infinity norm of a transfer function to the $\ell_2$-norm of its input and output signals.}
        
        To facilitate our discussion, we note that the relative degree of an LTI system $G(s) = \frac{n(s)}{d(s)}$ is $r = \text{deg}(d(s)) - \text{deg}(n(s))$, where $n(s)$ and $d(s)$ are polynomials in $s$, and $\text{deg}(\cdot)$ is the degree of a polynomial. We denote the relative degree of the source system $G_s(s)$ and target system $G_t(s)$ by $r_s$ and $r_t$.

        \begin{theorem}
        \label{theorem:upper_bound}
            Consider the control architecture in Fig.~\ref{subfig:block-diagram}, where an inverse module from robot $R_s$ is pre-cascaded to a target robot $R_t$. Under the assumptions \ref{assumption:min-phase}--\ref{assumption:yd-bounded-2norm}, and suppose that ${r_t \geq r_s}$, then the $\ell_2$-norm of the tracking error \diffht{of $R_t$ using an inverse module from $R_s$ can be bounded by}
            \begin{equation}
                \twonorm{e_{t,s}} \leq  \infnorm{E_{t,s}} \twonorm{y_d},
                \label{eq:e-2norm-ub}
            \end{equation}
            where $E_{t,s}(s)\mathord{=}\frac{\hat{e}_{t,s}(s)}{\hat{y}_d(s)}\mathord{=} G_s^{-1}(s) G_t(s)\mathord{-}1$ is the transfer function from the desired output $\hat{y}_d(s)$ to the target robot tracking error $\hat{e}_{t,s}(s)$, \diffhighlight{and $\infnorm{E_{t,s}} = \sup_{\omega\in\mathbb{R}} \abs{E_{t,s}\jw}$ is the infinity norm of $E_{t,s}$ with $\sup_{\omega\in\mathbb{R}}(\cdot)$ denoting the supremum of a function over frequencies $\omega\in\mathbb{R}$.}
        \end{theorem}

        Following the signal norm analysis in Sec.~2.3 of~\cite{doyle2013feedback}, we provide a proof of Theorem~\ref{theorem:upper_bound} below.

        \begin{proof}
          Consider source and target systems as defined by~\ref{assumption:linear-siso}. Following \ref{assumption:min-phase}, we assume that $G_s^{-1}(s)$ and $G_t(s)$ are BIBO stable. It follows that, if $r_t \geq r_s$, the error transfer function $E_{t,s}(s)=\frac{\hat{e}_{t,s}(s)}{\hat{y}_d(s)}$ is proper and BIBO stable. Since $E_{t,s}(s)$ is proper and stable, by Parseval's theorem~\cite{doyle2013feedback}, \diffhighlight{the $l_2$-norms of the error signal $e_{t,s}(t)$ and its Laplace transform $\hat{e}_{t,s}(s)$ are equal: $\twonorm{e_{t,s}} = \twonorm{\hat{e}_{t,s}}$. 
          Then, we obtain 
          \begin{equation*}
          \footnotesize
                \begin{split}
                    \twonorm{e_{t,s}}^2 = \twonorm{\hat{e}_{t,s}}^2
                    &= \frac{1}{2\pi}\int_{-\infty}^\infty \abs{E_{t,s}\jw}^2 \abs{\hat{y}_d\jw}^2 d\omega \\
                    &\leq \infnorm{E_{t,s}}^2\frac{1}{2\pi}\int_{-\infty}^\infty \abs{\hat{y}_d\jw}^2 d\omega \\
                    &= \infnorm{E_{t,s}}^2 \twonorm{y_d}^2,
                \end{split}
          \end{equation*}
        which leads to~\eqref{eq:e-2norm-ub} by taking the square root of each side.} Since $\twonorm{y_d}$ is finite by \ref{assumption:yd-bounded-2norm} and $E_{t,s}(s)$ is proper and stable, the upper bound given by $\infnorm{E_{t,s}}\twonorm{y_d}$ is finite. 
        \end{proof}

        \begin{remark}
            Given the baseline tracking error of the target system $e_{t,b}(t)$, we can guarantee positive transfer using the inverse module from system $R_s$ if ${\infnorm{E_{t,s}} \twonorm{y_d} < \twonorm{e_{t,b}}}$.
        \end{remark}
        
        Since $y_d$ is known, we can compute $\twonorm{y_d}$. To compute the upper bound in \eqref{eq:e-2norm-ub}, it remains to compute $\infnorm{E_{t,s}}$. In Sec.~\ref{subsec:bo-algorithm}, we propose an algorithm to estimate $\infnorm{E_{t,s}}$ for source and target systems with unknown dynamics.
    \subsection{Computing Tracking Error Bounds for Systems with Unknown Dynamics}
        \label{subsec:bo-algorithm}
        
        In this subsection, we assume that the exact dynamics of the source and the target systems are unknown and introduce an algorithm to estimate $\infnorm{E_{t,s}}$ from data. There exist methods in the literature to estimate the infinity norm of a transfer function from its input/output data, e.g. \cite{van2007data,doyle2013feedback}. However, our goal is to find an upper bound on $\twonorm{e_{t,s}}$ \textit{before} testing the transfer architecture in Fig. \ref{subfig:block-diagram}. As a result, we are unable to directly gather input/output data from $E_{t,s}$, and cannot use the aforementioned approaches in the literature. 
        
        To this end, we introduce a Bayesian Optimization (BO)-based algorithm to estimate the upper bound on $\twonorm{e_{t,s}}$ by running simple periodic experiments on the \textit{baseline systems} of the source and the target robots. Since we consider stable closed-loop baseline source and target systems, these simple periodic experiments can be realized over a practical set of frequencies of interest $\mathcal{W} =\{\omega\in\mathbb{R}\:\:|\:\: \omega_{\text{min}} \leq \omega \leq \omega_{\text{max}} \}$.
        
        To start our discussion, we write the estimation of $\infnorm{E_{t,s}}$ as the following optimization problem:
        \begin{equation}
            E_{t,s}^* = \sup_{\omega\in\mathcal{W}} f_{t,s}(\omega),
            \label{eq:max-E-approximation}
        \end{equation}
        where $E_{t,s}^*$ is an estimate of $\infnorm{E_{t,s}}$ \diffhighlight{and $f_{t,s}(\omega) = \abs{G_s^{-1}\jw G_t\jw - 1}$ is the objective function to be maximized.} Since $G_s$ and $G_t$ are assumed to be unknown, the objective function $f_{t,s}$ is unknown. To solve the black-box optimization problem in \eqref{eq:max-E-approximation}, we adopt a BO approach. In particular, we use a Gaussian Process (GP) model to approximate the objective function $f_{t,s}$ and optimize the objective function by iteratively sampling from $f_{t,s}$~\cite{shahriari2015taking}. 
        
        The algorithm runs as follows: We start with an initial $\omegasample$ that is randomly sampled over $\mathcal{W}$. Then, in each iteration, we proceed with the following steps: 
        
         \indent 1) \textit{Evaluating the unknown objective function:} We evaluate $f_{t,s}$ at $\omegasample$ by sending a sinusoidal reference with frequency $\omegasample$ to the source and target \textit{baseline} systems. From the magnitude~$M$ and phase~$\theta$ \diffht{of the baseline system input/output responses,} we can estimate $G_s^{-1}(j\omegasample)$ by $ M_s^{-1}\exp(-j\theta_s)$ and $G_t(j\omegasample)$ by $M_t\exp(j\theta_t)$, where subscripts $s$ and $t$ denote the source and target systems.
         The estimate of $f_{t,s}(\omegasample)$ can be computed from $G_s^{-1}(j\omegasample)$ and $G_t(j\omegasample)$. The \diffht{data} $(\omegasample, f_{t,s}(\omegasample))$ is added to a dataset $\mathcal{D} \mathord{=} \{(\omega_k, f_{t,s}(\omega_k))\}_{k=1}^{K}$, where \diffht{$K$ is the dataset size} at the current iteration.
        
         \indent 2) \textit{Fitting a GP model:} Given the dataset $\mathcal{D} = \{(\omega_k, f_{t,s}(\omega_k))\}_{k=1}^{K}$, we use a GP model to approximate $f_{t,s}$. The GP model has a prior mean function $\mu(\omega)$, and a kernel function $\kappa(\omega, \omega')$. 
         The GP posterior mean $\hat{f}_{t,s}(\omega)$ and posterior covariance $\hat{\sigma}_{t,s}^2(\omega)$ at a test point $\omega^*$ are 
        \begin{equation}
            \begin{split}
                \hat{f}_{t,s}(\omega^*) &= \mu(\omega^*) + (\bm{k}^{*})^T (\bm{K} + \sigma_n^2 I)^{-1}(\bm{f} - \bm{\mu}) \\
                \hat{\sigma}_{t,s}^2(\omega^*) &= \kappa(\omega^*, \omega^*) - (\bm{k}^{*})^T (\bm{K} + \sigma_n^2 I)^{-1}\bm{k}^*,
            \end{split}
        \label{eq:gp-posterior-mean-var}
        \end{equation}
        where $\bm{K}$ is the covariance matrix with the entry in the $i$th row and $j$th column given by  $[\bm{K}]_{ij} = \kappa(\omega_i, \omega_j)$, $\bm{k}^* = [\kappa(\omega^*, \omega_1), \dots, \kappa(\omega^*, \omega_K)]^T$, $\bm{f} = [f_{t,s}(\omega_1), \dots, f_{t,s}(\omega_K)]^T$, $\bm{\mu}= [\mu(\omega_1),\dots,\mu(\omega_K)]^T$, $\sigma_n^2$ is the noise variance, and $I$ is the identity matrix~\cite{shahriari2015taking}. 
        
        \indent 3) \textit{Determining the next sample to acquire:} In each iteration, we select the next sample $\omegasample$ based on an acquisition function. A common BO acquisition function is the expected improvement (EI), which is given by 
            \begin{equation}
                \text{EI}_{t,s}(\omega) = (\hat{f}_{t,s}(\omega) - f^\text{max}_{t,s}) \Phi(Z) + \hat{\sigma}_{t,s}(\omega)\phi(Z),
                \label{eq:EI}
            \end{equation}
            where $f^\text{max}_{t,s} = \max_k f_{t,s}(\omega_k)$ is the largest sample we have obtained thus far, ${Z = (\hat{f}_{t,s}(\omega) - f^\text{max}_{t,s}) / \hat{\sigma}_{t,s}(\omega)}$, and $\Phi(\cdot)$ and $\phi(\cdot)$ are the cumulative distribution and probability density functions of the standard normal distribution~\cite{shahriari2015taking}. The next point we sample at, $\omegasample$, is the one that maximizes the expected improvement: ${\omegasample = \argmax_{\omega \in \mathcal{W}} \text{EI}_{t,s}(\omega)}$. 
            
        \indent We repeat steps 1) -- 3) until the maximum of the GP posterior mean $\hat{f}_{t,s}$ converges to an approximately constant value. Our estimate of $\infnorm{E_{t,s}}$ is given by $\hat{E}^*_{t,s} = \hat{f}_{t,s}(\omega^*_{t,s}) + 3\hat{\sigma}_{t,s}(\omega^*_{t,s})$, where $\omega^*_{t,s} = \argmax_{\omega\in\mathcal{W}} \hat{f}_{t,s}(\omega)$. The corresponding estimate of the upper bound on $\twonorm{e_{t,s}}$ on a desired trajectory $y_d$ is $\Bar{e}_{t,s} = \hat{E}^*_{t,s} \twonorm{y_d}$.

        Note that we can extend the algorithm to the case when we have $N$ source robots $\{R_{s_n}\}_{n=1}^N$, and aim to estimate the corresponding upper bounds on $\twonorm{e_{t,s_n}}$ (${n = 1,\dots,N}$). In this case, we have $N$ optimization problems in the form of \eqref{eq:max-E-approximation}: ${E^*_{t,s_n} = \sup_{\omega \in \mathcal{W}} f_{t,s_n}(\omega)}$, where ${f_{t,s_n}(\omega) = \abs{G_{s_n}^{-1}\jw G_t\jw - 1}  }$. To solve them, we follow a similar approach to steps 1) -- 3): we model each objective function $f_{t,s_n}(\omega)$ with its own GP, with posterior mean $\hat{f}_{t,s_n}(\omega)$ and posterior variance $\hat{\sigma}^2_{t,s_n}(\omega)$. We replace the acquisition function in~\eqref{eq:EI} with $\alpha(\omega) = \max_n \text{EI}_{t,s_n}(\omega)$.
        The next sample point is the one that maximizes $\alpha(\omega)$: ${\omegasample = \argmax_{\omega \in \mathcal{W}} \alpha(\omega)}$.
        That is, given the expected improvement $\text{EI}_{t,s_n}(\omega)$ for each system, we sample at the point that gives the highest EI. Upon convergence, our estimate of $\infnorm{E_{t,s_n}}$ is given by $\hat{E}^*_{t,s_n} = \hat{f}_{t,s_n}(\omega^*_{t,s_n}) + 3\hat{\sigma}_{t,s_n}(\omega^*_{t,s_n})$, where $\omega^*_{t,s_n} = \argmax_{\omega\in\mathcal{W}} \hat{f}_{t,s_n}(\omega)$. The upper bound on $\twonorm{e_{t,s_n}}$ is given by $\Bar{e}_{t,s_n} = \hat{E}^*_{t,s_n} \twonorm{y_d}$. We summarize the overall algorithm in Alg.~\ref{alg:max-E-estimation}, and demonstrate the algorithm with quadrotor experiments in Sec.~\ref{sec:experiment}. \diffhighlight{Note that while we assume the source and target systems are linear, Alg.~\ref{alg:max-E-estimation} uses only their input/output data, without requiring knowledge of the system structure (e.g., number of poles or zeros)}.
    
        \SetAlCapNameFnt{\footnotesize}
        \SetAlCapFnt{\footnotesize}
        \begin{algorithm}[!t]
            \footnotesize
            \Input{ $N$ source robots $\{R_{s_n}\}_{n=1}^N$, one target robot $R_t$, desired trajectory $y_d$, and a frequency range $\mathcal{W} =\{\omega\in\mathbb{R}\:\:|\:\: \omega_{\text{min}} \leq \omega \leq \omega_{\text{max}} \}$. }
            \Output { Upper bound $\Bar{e}_{t,s_n}$ on $\twonorm{e_{t,s_n}}$.}
            \Init{Empty sample datasets: $\mathcal{D}_n \gets \varnothing \ (n=1,\dots,N)$} 
            $ \text{Compute initial sample: } \omegasample \sim \mathcal{U}(\omega_\text{min}, \omega_\text{max})$\\
            \While{not converged}
                {
                    Estimate $G_t(j\omegasample)$\\
                    \For{$n = 1, \dots, N$}
                    {
                        Estimate $G_{s_n}^{-1}(j\omegasample)$ \\
                        
                        Compute $f_{t,s_n}(\omegasample)$ \\
                        
                        $\mathcal{D}_n \gets \mathcal{D}_n \cup \{(\omegasample, f_{t,s_n}(\omegasample))\}$
                        
                        Fit $n$th GP with data $\mathcal{D}_n$
                        
                    } 
                    Compute $\omegasample = \argmax_{\omega \in \mathcal{W}} \alpha(\omega)$
                } 
            
            \For{$n = 1, \dots, N$}
            {
                $\omega^*_{t,s_n} = \argmax_{\omega\in\mathcal{W}} \hat{f}_{t,s_n}(\omega)$
                
                $\hat{E}^*_{t,s_n} = \hat{f}_{t,s_n}(\omega^*_{t,s_n}) + 3\hat{\sigma}_{t,s_n}(\omega^*_{t,s_n})$ \\
                
                $\Bar{e}_{t,s_n} = \hat{E}^*_{t,s_n} \twonorm{y_d}$

            }
            \caption{ Upper Bound on~$\twonorm{e_{t,s_n}}$}
            \label{alg:max-E-estimation}
        \end{algorithm}

    \subsection{Asymmetry in Inverse Module Transfer Learning}
        \label{subsec:asymmetry}
    
        In this section, we explore the asymmetric nature that is inherent in transfer learning. In the following discussion, we first demonstrate the asymmetry property for two first-order systems in the inverse transfer learning framework and then extend our discussion to a more generalized setting. 
        
        We consider two first-order systems $G_1(s)=\frac{1}{\tau_1 s + 1}$ and $G_2(s)= \frac{1}{\tau_2 s + 1}$, where $\tau_1,\ \tau_2 > 0$ are the time constants of each system. The baseline tracking error of each system is denoted $e_{1,b}(t)$ and $e_{2,b}(t)$, respectively. The tracking error when pre-cascading $G_1^{-1}$ to $G_2$ and when pre-cascading $G_2^{-1}$ to $G_1$ are denoted by $e_{2,1}(t)$ and $e_{1,2}(t)$, respectively. 
        \begin{example}
            Consider the two first-order systems with transfer functions $G_1(s)$ and $G_2(s)$. Without loss of generality, we assume $\tau_2 < \tau_1$. Then, for a given desired trajectory $y_d$, transferring the inverse $G_2^{-1}$ to $G_1$ always leads to positive transfer, i.e. $\twonorm{e_{1,2}} < \twonorm{e_{1,b}}$. However, the reverse is not true. \diffht{For} the same trajectory $y_d$, transferring the inverse $G_1^{-1}$ to $G_2$ is positive ($\twonorm{e_{2,1}} < \twonorm{e_{2,b}}$) iff $\tau_1$ satisfies $\tau_1 < 2\tau_2$.
             
            \label{ex:asymmetry-fos}
        \end{example}
        
        \begin{proof}
            By directly evaluating the convolution integral for the output of a linear system, one can show that $e_{1,b}(t) = \frac{1}{\tau_1}e^{-t/\tau_1} \mathcal{I} - y_d(t)$ and $e_{1,2}(t) = \left(1 - \frac{\tau_2}{\tau_1} \right)e_{1,b}(t)$, where $\mathcal{I} = \int_0^t y_d(z) e^{z/\tau_1} dz$. It follows that ${ \twonorm{ e_{1,2} } < \twonorm{ e_{1,b} } }$ iff ${\tau_2 < 2\tau_1}$. By assumption, $\tau_2< \tau_1$, and thus the condition of positive transfer from $G_2^{-1}$ to $G_1$ is always satisfied. For the reverse direction, one can similarly show that  ${ \twonorm{ e_{2,1} } < \twonorm{ e_{2,b} } }$ iff ${\tau_1 < 2\tau_2}$. Thus, transfer from $G_1^{-1}$ to $G_2$ is positive iff $\tau_1$ is sufficiently small (i.e., ${\tau_1 < 2\tau_2}$). 
        \end{proof}
        
        \begin{remark}
           For the systems we consider in Example~\ref{ex:asymmetry-fos}, the time constant $\tau$ of each system represents the rise time of the system in response to a step input, which can be interpreted as the system's aggressiveness with a smaller value indicating a higher aggressiveness.
           The result then implies that it is easier to guarantee positive transfer from a more aggressive system to a less aggressive system, but not vice versa. 
        \end{remark}

        We can extend the insight by leveraging the $\nu$-gap metric from robust control, which provides a notion of `distance' between two systems. For two SISO minimum phase systems satisfying $\infnorm{G_2(-j\omega)G_1\jw)}\mathord{<}1$, the $\nu$-gap metric can be written as $\delta_\nu(G_1, G_2) = \sup_\omega\psi(G_1\jw,G_2\jw)$, where $\psi(G_1\jw,G_2\jw) =  \frac{\abs{G_1\jw - G_2\jw}}{\sqrt{(1+\abs{G_1\jw}^2)(1+\abs{G_2\jw}^2)}}$
        is the chordal distance between the projections of $G_1\jw$ and $G_2\jw$ onto the Riemann sphere~\cite{zhou1998essentials}.
        For a desired trajectory with frequency $\omega$, we can decompose the error transfer function into a symmetric and an asymmetric component:
        \begin{align*}
            \footnotesize
            \abs{E_{1,2}\jw} = \underbrace{\psi(G_1\jw,G_2\jw)}_{\text{symmetric}}\underbrace{\Psi(G_1\jw,G_2\jw)}_{\text{asymmetric}},
        \end{align*}
        where $\Psi(G_1\jw,G_2\jw)=\frac{\sqrt{1+\abs{G_1\jw}^2}}{\psi(G_2\jw,0)}$.
        The term $\psi(G_1\jw,G_2\jw)$ is the symmetric term providing a notion of `distance' between $G_1$ and $G_2$. Systems close in this sense will have reduced $\abs{E_{1,2}\jw}$. The term $\Psi(G_1\jw,G_2\jw)$ reflects the direction of transfer. Note that, since $\psi(G_2\jw,0)$ represents the chordal distance between $G_2\jw$ and $0$, we can interpret $\psi(G_2\jw,0)$ as a measure of the aggressiveness of the system. As a result, if $G_2$ is more aggressive, then $\Psi(G_1\jw,G_2\jw)$ and thus $\abs{E_{1,2}\jw}$ is further reduced. This implies that transferring an inverse module from a more aggressive system is more likely to reduce the tracking error of a less aggressive system and thereby results in positive transfer.
        
        We note that the asymmetric nature of transfer learning is not limited to our inverse module transfer framework. The asymmetric behaviour is also found in frameworks such as alignment-based transfer learning~\cite{helwa2017multi,raimalwala2015upper}. To guarantee positive transfer, it is important to define a measure of similarity between the source and target domains~\cite{rosenstein2005transfer,wang2019characterizing}. Such measures of similarity should account for the direction of transfer, which has a direct impact on transfer success.

\section{Quadrotor Experiments}
In this section, we demonstrate Alg.~\ref{alg:max-E-estimation} and the asymmetric nature of the transfer problem with quadrotor experiments.
    \label{sec:experiment}
    \subsection{Experiment Setup}
        We consider one target robot $R_t$, whose tracking is to be improved, and five source robots (Fig.~\ref{subfig:source-target-quadrotors}), whose inverse modules have been previously trained and can be leveraged to improve the tracking of $R_t$. \diffhighlight{The target robot in our experiments is the Crazyflie 2.0 quadrotor; its mass is 36~g, with a rotor-to-rotor distance of 14~cm.}  
        The five source robots are: $R_{s_1}$~-- Crazyflie 2.0 quadrotor (light), $R_{s_2}$~-- Crazyflie simulator, $R_{s_3}$~-- Parrot ARDrone 2.0, $R_{s_4}$~-- Parrot ARDrone simulator, and $R_{s_5}$~-- Parrot Bebop 2. \diffhighlight{The simulators $R_{s_2}$, $R_{s_4}$ emulate the dynamics and control of $R_{t}$ and $R_{s_3}$, but do not account for effects such as aerodynamic drag or time delays.
        The mass of the source robots are 0.86, 1, 13.7, 13.7 and 14.5 times that of the target, respectively, and the rotor-to-rotor distance of the source robots are 1, 1, 3.9, 3.9, and 3.1 times that of the target, respectively}.
        \diffhighlight{The baseline controllers for the target $R_t$ and sources $R_{s_1}$ and $R_{s_2}$ are designed by exploiting the differential flatness of quadrotors~\cite{mellinger2011minimum}, while the baseline controllers for sources $R_{s_3}$--$R_{s_5}$ are designed based on a combination of feedback linearization and PD control~\cite{helwa2018construction}.} We assume that the position dynamics of each quadrotor is decoupled in the $x$-, $y$-, and $z$-directions. 
        The inverse module of each source system is approximated by a deep neural network (DNN) trained on 400 seconds of data collected on each source baseline system~\cite{zhou-cdc17}.  
        
        Our experiments consist of three parts: in Sec.~\ref{subsec:experiments-ub-estimation}, we apply Alg.~\ref{alg:max-E-estimation} to estimate the upper bound on tracking error of the target quadrotor when using the inverse module from each source quadrotor. In Sec.~\ref{subsec:experiments-tracking-tests}, we verify the upper bounds on five random periodic trajectories. In Sec.~\ref{subsec:experiments-asymmetry}, we illustrate the asymmetric nature of this transfer problem. \diffhighlight{We note that in the upper bound estimation experiments and test trajectory experiments, we rely on the input/output data from the source and target quadrotors, and do not require knowledge of their exact dynamics or system structures.}

    \subsection{Tracking Error Upper Bound Estimation}
        \label{subsec:experiments-ub-estimation}
            
        We apply Alg.~\ref{alg:max-E-estimation} proposed in Sec.~\ref{subsec:bo-algorithm} to identify the upper bounds on the tracking errors of the target quadrotor system pre-cascaded with the DNN inverse modules of the source quadrotors. Since by assumption the dynamics of the source and target systems are decoupled in the $x$-, $y$-, and $z$-directions, we apply Alg.~\ref{alg:max-E-estimation} to each direction separately. We denote the tracking error of $R_t$ using an inverse module from source $R_{s_n}$ in the $x$-, $y$-, and $z$-directions by ${e_{t,s_n, x }(t)}$, ${e_{t,s_n, y}(t)}$, and ${e_{t,s_n, z }(t)}$, respectively. Our goal is to estimate $\hat{E}^*_{t,s_n,x}$, $\hat{E}^*_{t,s_n,y}$, and $\hat{E}^*_{t,s_n,z}$ to compute the error bounds on $\twonorm{e_{t,s_n, x }}$, $\twonorm{e_{t,s_n, y }}$, and $\twonorm{e_{t,s_n, z }}$ (see~\eqref{eq:e-2norm-ub}).  
        As outlined in Alg.~\ref{alg:max-E-estimation}, we estimate the error norms $\hat{E}^*_{t,s_n,x}$, $\hat{E}^*_{t,s_n,y}$, and $\hat{E}^*_{t,s_n,z}$ iteratively by running periodic trajectories on the baseline systems along the $x$-, $y$-, and $z$-directions. The estimates of the error norms converged in six iterations in the $x$- and $y$-directions, and five iterations in the $z$-direction. 
        
        The converged estimates $\hat{E}^*_{t,s_n,x}$, $\hat{E}^*_{t,s_n,y}$, and $\hat{E}^*_{t,s_n,z}$ are summarized in Table~\ref{table:E-estimates}. \diffhighlight{Sources $R_{s_1}$ and $R_{s_2}$ are of similar size and agility as the target $R_t$, and thus the upper bound on tracking error in each direction when using their inverse modules is low. Conversely, sources $R_{s_3}$--$R_{s_5}$ are much larger and less agile than the target, leading to a larger upper bound on tracking error when using their inverse modules. These differences are especially pronounced in the $x$- and $y$-directions. 
        \diffht{From} Table~\ref{table:E-estimates}, we expect transferring the inverse modules from source quadrotors similarly aggressive to the target ($R_{s_1}$ and $R_{s_2}$) would lead to lower tracking error on the target quadrotor as compared to transferring the inverse modules from less aggressive source quadrotors ($R_{s_3}$--$R_{s_5}$).
        } 
        
        \diffhighlight{Note that to compute the estimates in Table~\ref{table:E-estimates}, we required 209 seconds of data from the target quadrotor, approximately half of the data needed to train its inverse module from scratch.}
        As we show in Sec.~\ref{subsec:experiments-tracking-tests}, the estimated upper bounds allow us to safely reuse an inverse from a source quadrotor to improve the target quadrotor tracking performance. 
        
        \begin{table}[]
        \vspace{0.5em}
        \caption{A summary of the error norm estimates $\hat{E}^*_{t,s_n}$ of the target quadrotor when using the inverse modules from the five source quadrotors.}
        \vspace{0.5em}
            \centering
            \begin{tabular}{c|l|l|l|l|l}
            \hline\hline
                \rule{0pt}{1.1em}
                Direction & $\hat{E}^*_{t,s_1}$ & $\hat{E}^*_{t,s_2}$ & $\hat{E}^*_{t,s_3}$ & $\hat{E}^*_{t,s_4}$ & $\hat{E}^*_{t,s_5}$ \\[0.5ex]
                \hline
            $x$ & 0.22      & 0.25      & 2.70      & 2.19      & 3.03      \\
            $y$ & 0.23      & 0.28      & 3.20      & 2.23      & 3.25      \\
            $z$ & 0.22      & 0.24      & 0.56      & 1.55      & 0.98 \\\hline\hline
            \end{tabular}
            \label{table:E-estimates}
        \end{table}

    \subsection{Test Trajectory Tracking Experiments}
        \label{subsec:experiments-tracking-tests}
        
        \diffhighlight{We validate the estimated upper bounds on the target system tracking error (Table~\ref{table:E-estimates}) on five test trajectories. The test trajectories are parameterized as $y_{d}(t) = [0.25\sin{(\omega_x t)},\: 0.25\cos{(\omega_y t)},\: 0.25\sin{(\omega_z t)}]$, where $\omega_x$, $\omega_y,$ and $\omega_z$ are randomly sampled between $0$ and $2$~rad/sec, which is the maximal range for the target quadrotor to track.
        } 
        
        \diffhighlight{We quantify the tracking performance of the target quadrotor as $e_{t,\cdot} = (\twonorm{e_{t,\cdot,x}}^2+\twonorm{e_{t,\cdot,y}}^2+\twonorm{e_{t,\cdot,z}}^2)^{1/2}$. Using the estimates in Table~\ref{table:E-estimates}, we can bound the tracking error of the target $R_t$ when using the inverse module from~$R_{s_n}$ as ${e_{t,s_n} \leq e^*_{t,s_n}}$, where $e^*_{t,s_n} = ((\hat{E}^*_{t,s_n,x}\twonorm{y_{d,x}})^2 +(\hat{E}^*_{t,s_n,y}\twonorm{y_{d,y}})^2 +(\hat{E}^*_{t,s_n,z}\twonorm{y_{d,z}})^2)^{1/2}$}. \diffhighlight{We guarantee positive transfer from $R_{s_n}$ to $R_{t}$ if the condition ${e^*_{t,s_n} < e_{t,b}}$ is satisfied.}
        
        \diffhighlight{
        We ran each test trajectory on the target quadrotor baseline system with and without the source inverse modules. As shown in Table~\ref{tab:tracking_results}, for each trajectory and each source robot $R_{s_n}$, we have that $e_{t, s_n} \mathord{<} e^*_{t,s_n}$, verifying the efficacy of the upper bounds. By comparing the baseline tracking error $e_{t,b}$ to the upper bounds $e^*_{t, s_n}$, we observe that we can guarantee positive transfer for sources $R_{s_1}$ and $R_{s_2}$, as $e^*_{t, s_n} \mathord{<} e_{t,b}$ for $n\mathord{\in}\{1,2\}$. \diffht{From} the estimated upper bounds, we can predict positive transfer for $R_{s_1}$ and $R_{s_2}$ before testing their inverse modules on the target quadrotor. Using inverse modules from $R_{s_1}$ and $R_{s_2}$ improve the target quadrotor performance by an average of 74\% without requiring retraining an inverse module for the target system. Conversely, for $R_{s_3}$--$R_{s_5}$, we cannot guarantee positive transfer; indeed, when using these source inverse modules, we observe negative transfer ($e_{t, s_n} \mathord{>} e_{t,b}$). 
        }

\begin{table*}[]
\vspace{0.5em}
\centering
\caption{\diffhighlight{Quadrotor trajectory tracking experiment results. For each source system $R_{s_n}$, we display the actual tracking error $e_{t,s_n}$, the tracking error upper bound $e^*_{t,s_n}$, and the baseline tracking error $e_{t,b}$ of the target robot. Positive transfer cases (i.e., entries with values less than $e_{t, b}$) are indicated with boldface. For all cases, we have that ${e_{t, s_n} < e^*_{t,s_n}}$, verifying the efficacy of the upper bounds. Positive transfer is guaranteed for $R_{s_1}$ and $R_{s_2}$. 
}}
\diffhighlight{
  \begin{tabular}{c|SS|SS|SS|SS|SS|c}
    \toprule
    \multirow{2}{*}{Traj.$^\dagger$} &
      \multicolumn{2}{c|}{Transfer from $R_{s_1}$} &
      \multicolumn{2}{c|}{Transfer from $R_{s_2}$} &
      \multicolumn{2}{c|}{Transfer from $R_{s_3}$} &
      \multicolumn{2}{c|}{Transfer from $R_{s_4}$} &
      \multicolumn{2}{c|}{Transfer from $R_{s_5}$} &
      \multicolumn{1}{c}{Baseline} \\
       &  {$e_{t,s_1}$[m]} & {$e^*_{t,s_1}$ [m]} & {$e_{t,s_2}$ [m]} & {$e^*_{t,s_2}$ [m]} & {$e_{t,s_3}$ [m]} & {$e^*_{t,s_3}$ [m]} & {$e_{t,s_4}$ [m]} & {$e^*_{t,s_4}$ [m]} & {$e_{t,s_5}$ [m]} & {$e^*_{t,s_5}$ [m]} & {$e_{t,b}$ [m]} \\
      \midrule
    $y_{d_1}$ & \hspace{1.45em}\textbf{0.144} & \hspace{1.45em}\textbf{0.358} & \hspace{1.45em}\textbf{0.161} & \hspace{1.45em}\textbf{0.335} & 0.775 & 2.401 & \hspace{1.45em}\textbf{0.542} & 2.440 & 0.610 & 3.263 & 0.562   \\
    $y_{d_2}$ & \hspace{1.45em}\textbf{0.174} & \hspace{1.45em}\textbf{0.419} & \hspace{1.45em}\textbf{0.162} & \hspace{1.45em}\textbf{0.399} & 0.834 & 2.990 & \hspace{1.45em}\textbf{0.557} & 2.871 & 0.822 & 3.962 & 0.665   \\
    $y_{d_3}$ & \hspace{1.45em}\textbf{0.224} & \hspace{1.45em}\textbf{0.358} & \hspace{1.45em}\textbf{0.210} & \hspace{1.45em}\textbf{0.335} & 0.926 & 2.396 & 1.041 & 2.441 & 0.923 & 3.264 & 0.817   \\
    $y_{d_4}$ & \hspace{1.45em}\textbf{0.215} & \hspace{1.45em}\textbf{0.354} & \hspace{1.45em}\textbf{0.215} & \hspace{1.45em}\textbf{0.331} & 0.890 & 2.382 & 1.076 & 2.418 & 0.903 & 3.244 & 0.867   \\
    $y_{d_5}$ & \hspace{1.45em}\textbf{0.252} & \hspace{1.45em}\textbf{0.357} & \hspace{1.45em}\textbf{0.258} & \hspace{1.45em}\textbf{0.334} & \hspace{1.45em}\textbf{0.945} & 2.389 & 1.110 & 2.434 & 1.081 & 3.252 & 0.972   \\
    \bottomrule
  \end{tabular}
  \vspace{0.2em}
  \caption*{\scriptsize
  $^\dagger$The frequencies $(\omega_x,\omega_y,\omega_z)$ of the five sinusoidal test trajectories were uniformly sampled between 0 and 2 rad/sec using the Matlab function \texttt{rand}. The values are $(\omega_x,\omega_y,\omega_z)\in\{(1.087,0.557,0.849), (1.690,0.009,0.243), (1.342,1.652,0.273), (1.150,1.783,0.418), (1.957,1.623,0.344) \}$ rad/sec.}
}
\label{tab:tracking_results}
\vspace{-1.5em}
\end{table*}

    \subsection{The Asymmetric Nature of Inverse Module Transfer}
        \label{subsec:experiments-asymmetry}
        Although from the tracking experiments, we see that using an inverse module from a less similar robot (e.g., $R_{s_5}$) can cause negative transfer to the target $R_t$, the reverse is not necessarily true.  To demonstrate the asymmetric nature of transfer, we train an inverse module for the target robot $R_t$ and pre-cascade this module to the baseline controller of $R_{s_5}$. 
        
\begin{figure}
            \includegraphics[width=0.9\columnwidth,keepaspectratio]{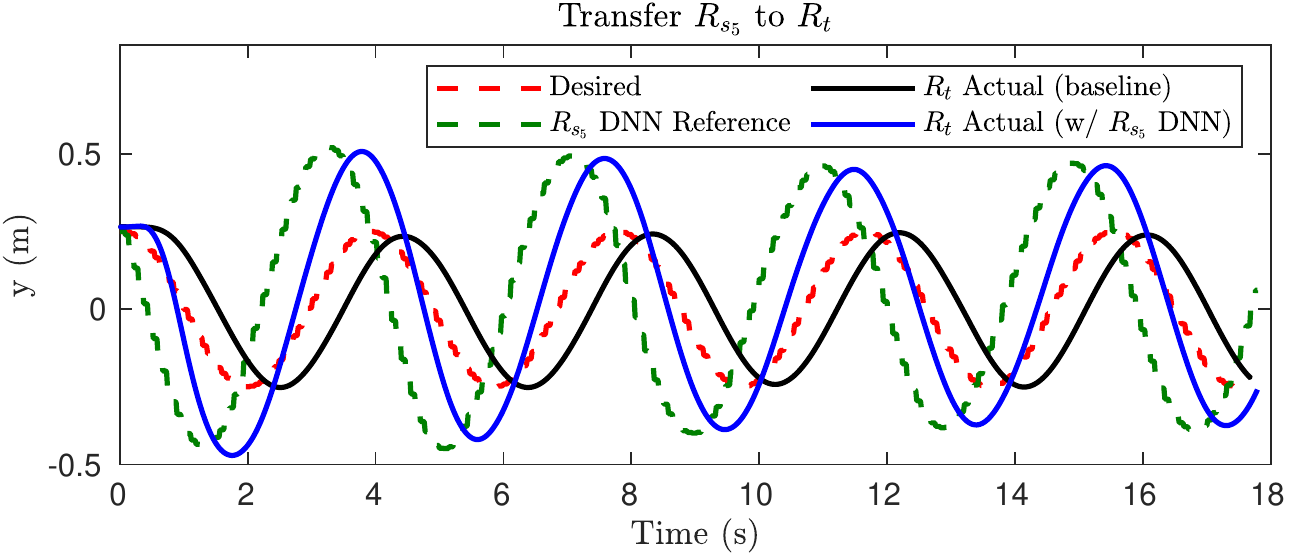}
            \caption{Using an inverse module from $R_{s_5}$ for $R_t$ increases the tracking error of $R_t$, ${\twonorm{e_{t,s_5,y}}}$, by 13\%, relative to its baseline tracking error. The inverse module from $R_{s_5}$ amplifies the reference $y_d$ (red) to produce a reference $y_r$ (green). The over-amplification of the reference adversely causes an increase in tracking error of $R_t$ compared to its baseline (black). }
            \label{fig:asymmetry-experiment-example}
        \end{figure}

        As an example, in the $y$-direction for one test trajectory from Sec.~\ref{subsec:experiments-tracking-tests}, when using the inverse module from $R_{s_5}$, the tracking error of $R_t$, $\twonorm{e_{t,s_5,y}}$, is \textit{increased} by 13\% relative to its baseline error. 
        However, when transferring the inverse module from $R_t$ to $R_{s_5}$, the tracking error of $R_{s_5}$, $\twonorm{e_{s_5,t,y}}$, is \textit{reduced} by 9\%, relative to its baseline error. The intuition of this asymmetry is discussed in Sec.~\ref{subsec:asymmetry}. 
        The negative transfer from $R_{s_5}$ to $R_{t}$ is a result of $R_{s_5}$ being less aggressive. As compared to $R_{t}$, the baseline system of $R_{s_5}$ has larger delays and damping. To compensate for the slow response, as seen in Fig.~\ref{fig:asymmetry-experiment-example}, given a desired trajectory (red), the inverse of $R_{s_5}$ tends to produce a significantly amplified reference (green). When sending the amplified reference to the target robot $R_t$, the target robot $R_t$ follows the $R_{s_5}$ DNN reference (green) `too well', causing it to overshoot the desired trajectory. Conversely, when pre-cascading the inverse of $R_t$, the more aggressive robot, to $R_{s_5}$, the less aggressive robot, the $R_t$ inverse module slightly amplifies the reference signal to compensate for the slow response of $R_{s_5}$. This results in positive transfer from $R_t$ to $R_{s_5}$.

\section{Conclusion and Future Work}
    \label{sec:conclusion}
    
    In this paper, we considered an inverse module transfer framework in which an inverse dynamics module from a source system is transferred to a target system with the goal of improving the tracking performance of the target system. We derived an upper bound on the $l_2$-norm of the tracking error of the target system when using a source inverse module and proposed an algorithm to estimate this bound using input/output data from the source and target systems. We also highlighted the asymmetric nature of this transfer framework. In quadrotor experiments, we demonstrated the efficacy of the upper bounds, showed how the estimated bound can be used for guaranteeing positive transfer, and illustrated the asymmetric nature in the inverse transfer framework with two quadrotors cross-sharing their inverse modules. 

    As future work, we would like to extend our theoretical results on transfer performance guarantees and the asymmetric nature of this transfer framework to multi-input multi-output nonlinear \diffht{systems} and explore the asymmetric nature of transfer learning in other robot learning-based control frameworks.

\bibliographystyle{IEEEtran}
\bibliography{IEEEabrv,references}

\end{document}